\documentclass[11pt]{article}
\usepackage{coling2020}
\usepackage{times}
\usepackage{url}
\usepackage{latexsym}

\newcommand{\comment}[1]{}

\usepackage{amssymb}
\usepackage{pifont}
\newcommand{\xmark}{\ding{55}}%

\usepackage{times}
\usepackage{latexsym}

\usepackage{amssymb}
\usepackage{pifont}

\usepackage{floatrow}
\newfloatcommand{capbtabbox}{table}[][\FBwidth]
\usepackage{blindtext}
\usepackage{amssymb} 
\usepackage{amsmath}

\usepackage{diagbox}
\usepackage{booktabs}
\usepackage{lipsum}
\usepackage{mwe}
\usepackage{multirow}
\usepackage{bigdelim}

\usepackage{caption}
\usepackage{subcaption}

\usepackage{array}
\usepackage{makecell}

\newcolumntype{g}{>{\columncolor{green}}c}

\usepackage[noend]{algpseudocode}

\usepackage[mathscr]{euscript}
 \let\mathscr\relax

\usepackage[inline]{enumitem}

\DeclareTextCommandDefault{\textcopyright}{\textcircled{c}}

\def\app#1#2{%
  \mathrel{%
    \setbox0=\hbox{$#1\sim$}%
    \setbox2=\hbox{%
      \rlap{\hbox{$#1\propto$}}%
      \lower1.1\ht0\box0%
    }%
    \raise0.25\ht2\box2%
  }%
}

\usepackage{url}

\usepackage{color}
\usepackage{cleveref}

\colingfinalcopy 

\title{DART: A Lightweight Quality-Suggestive Data-to-Text Annotation Tool}

\author{ 
  \dag Ernie Chang, \dag Jeriah Caplinger, *Alex Marin,  \dag Xiaoyu Shen, \dag Vera Demberg \\
  \dag Dept. of Language Science and Technology, Saarland University \\
    {\tt \{cychang,jeriahc,xiaoyu,vera\}@coli.uni-saarland.de}
  \\ *Microsoft Corporation, Redmond, WA
  \\ 
  {\tt \{alemari\}@microsoft.com}\\
}

\date{}

\begin{document}
\maketitle
\begin{abstract}
We present a lightweight annotation tool, the \emph{ Data AnnotatoR Tool (DART)}, for the general task of labeling structured data with textual descriptions. 
The tool is implemented as an interactive application that reduces human efforts in annotating large quantities of structured data, \emph{e.g.} in the format of a table or tree structure. 
By using a backend sequence-to-sequence model, our system iteratively analyzes the annotated labels in order to better sample unlabeled data.
In a simulation experiment performed on annotating large quantities of 
structured data, 
\emph{DART} has been shown to reduce the total number of annotations needed with active learning and automatically suggesting relevant labels.  

\end{abstract}

\section{Introduction}

\blfootnote{
    %
    %
    \hspace{-0.65cm}  
    This work is licensed under a Creative Commons Attribution 4.0 International License. License details: \url{http:// creativecommons.org/licenses/by/4.0/}
    %
    %
    %
    %
}

Neural data-to-text generation has been the subject of much research in recent years \cite{gkatzia2016content}. Traditionally, the task takes as input structured data which comes in the form of tables with attribute and value pairs, and generates free-form, human-readable text.
\comment{
\begin{figure}[h]
\centering

\parbox{0.9\linewidth}{
\textbf{Name}[Clowns], \\
\textbf{PriceRange}[more than \pounds 30], \\
\textbf{EatType}[pub], \\
\textbf{FamilyFriendly}[no]
}
    \caption{ Sample structured data containing attribute (\emph{e.g.} \textit{Name}) and value (\emph{e.g.} \textit{Clowns}) .
    }
    \label{fig:sample}
\end{figure}}
\begin{figure}[h]
  \centering
\includegraphics[width=0.85\columnwidth]{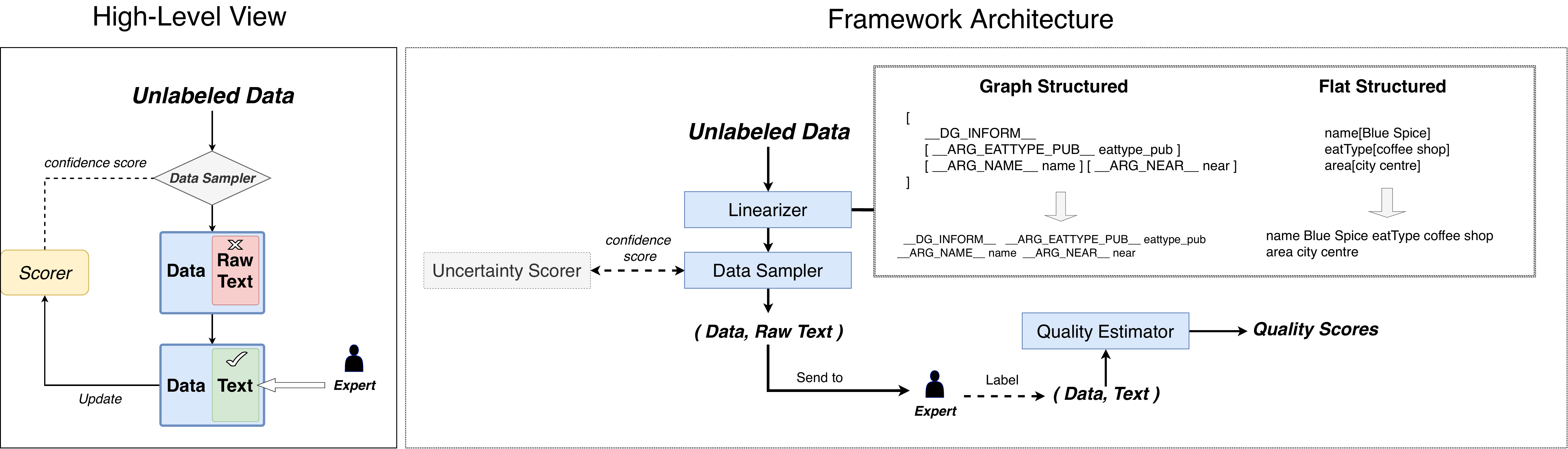}
\caption{ Left: Overview of the DART toolkit usage. Right: Diagram of the framework architecture. }
\label{fig:overview}
\end{figure}

Past example datasets include Restaurants \cite{wen-etal-2015-semantically} or graph-structure inputs \cite{constrained}.  
Analogously, most conversation systems \cite{luis,crook2018conversational} utilize intermediate meaning representation (data) as input to generate natural language sentences. 
In practice, however, these systems are highly reliant on the use of large-scale labeled data. Each new domain requires additional annotations to pair the new data with matching text. 
With the rise in development of natural language generation (NLG) systems from structured data, there is also an increased need for annotation tools that reduce labeling time and effort for constructing complex sentence labels.
Unlike other labeling tasks, such as sequence tagging \cite{lin-etal-2019-alpacatag}, where the labels are non-complex and correspond to fixed sets of classes, 
data-to-text generation entails providing complete sentence labels for each data instance. 
To construct textual description is time-consuming and therefore it can be beneficial for the system to automatically suggest texts and allow the annotators to accept or \emph{partially correct} them. 
To this end, we propose to create an interactive annotation tool: Data AnnotatoR Tool (DART\footnote{Demo is available at \url{https://youtu.be/onPUgQ2ixpI} }) that reduces structured data-to-text annotation efforts by incorporating 
\emph{automatic label suggestion} and the 
\emph{uncertainty-based active learning algorithm} \cite{lewis1994heterogeneous,culotta2004confidence}.
\emph{DART} serves as a natural complement to downstream data-to-text systems, rather than an end-to-end NLG system.
As such, it can assist in the development of both traditional rule-based systems (\emph{e.g.} \cite{reiter2007architecture}), and the recent neural systems (\emph{e.g.} \cite{constrained,chang2020unsupervised,shen2020neural,hong2019improving}).

\begin{table*}
  \centering
  \small
  \begin{tabular}{lccccc}
    \toprule
           & Label &    Programming & Has Label  &     User & Use Active \\
    System &      Types &  Language & Recommendation ? & Interface & Learning ? \\
    \midrule
    DART & Text  &  Python & \checkmark & GUI & \checkmark \\
    AlpacaTag \cite{lin-etal-2019-alpacatag} & Tag  &  Python & \checkmark & GUI & \checkmark \\
    YEDDA	\cite{yang2018yedda} & Tag  &  Python & \checkmark & GUI & \xmark \\
    \bottomrule
  \end{tabular}
  \caption{\label{tab:comparison}
  A general comparison of relevant GUI-based annotation tools.
  }
\end{table*}

As a lightweight, standalone desktop application, \textit{DART} can be easily distributed to domain experts and installed on local devices.
\textit{DART} consists of a user-friendly interface that allows experts to iteratively improve the overall corpus quality with partial corrections. 
Overall, the toolkit provides three advantages: 
(1) It reduces labeling difficulty by automatically providing natural language label recommendations;
(2) It efficiently solicits data for which it has low confidence (or high uncertainty) in its generated text to be annotated, so that overall annotation efforts can be reduced;
(3) Lastly, it provides real-time in-progress updates with statistics about the labeled corpus as to help direct the overall annotation process.
This is achieved with a myriad of quality estimators that assess corpus diversity and the overall text quality.

\section{Annotation Framework}
\label{sec:framework}

\emph{DART} is a desktop application built with PyQt5\footnote{\url{https://riverbankcomputing.com/software/pyqt/intro}}.
It is compiled into a single executable with PyInstaller\footnote{\url{https://www.pyinstaller.org/}}, a tool that supports both Mac OS and Windows environments. 
It contains an intuitive interface as described in \cref{sec:interface}.
Annotation experts interact with \emph{DART} in the following way:
(1) A file containing unlabeled data is uploaded.
(2) The system samples some data instances from the file, with a selection strategy based on signals from the sequence-to-sequence \emph{uncertainty scorer} (\cref{sec:cyclic}) and performed with the \emph{data sampler} (\cref{sec:select}).
(3) Experts then annotate the provided data by correcting the suggested labels (available after the first iteration of (1)-(2)).
(4) During the process of annotation, the labeled corpus quality is indicated by the annotation \emph{quality estimators} (\cref{sec:estimator}) for experts to determine if the process were to be terminated.
We discuss each component in more detail below.

\subsection{Uncertainty Scorer}
\label{sec:cyclic}

We represent the structured unannotated corpus as $D = d_{i=1}^{N}$ where each data sample $d_i$ comprises of a token sequence linearized from underlying structured data samples $x_i$, as motivated by past works in the multilingual surface realization tasks \cite{ws-2018-multilingual}. 
We employ the Transformer-based~\cite{vaswani2017attention} encoder-decoder architecture as the sequence-to-sequence model.
The sequences $d_i$ are fed into the model in order to generate a text sequence $t_i =  {w_1,w_2,...,w_{M_i}}$ of length $M_i$.

Since the model is given only the input data $d$, we compute reconstruction scores for this data, and use the cross-entropy loss as the \emph{uncertainty score}. 
To do so, we perform round-trip training\footnote{ Defined in \cite{lample2017unsupervised} as the back-translation technique where  input data is reconstructed with $M_{forward}$ and $M_{backward}$ to compute consistency loss.} where the source data is reconstructed to achieve \emph{cycle consistency}.
In this setup, the same encoder and decoder are used in both forward and backward training \emph{i.e.} a forward model $M_{forward}$ goes from data to text and the backward model $M_{backward}$ converts the text back into data.
We define the round-trip training log-loss as the uncertainty score $S_{uncertainty}$ (as Eq.~\ref{eq:forward}) where $\mathcal{L}( \cdot )$ is the cross-entropy loss.
$d'$ is the generated data using $M_{forward}$ and $M_{backward}$ given input data $d$.
\begin{subequations}
\begin{align}
    \mathcal{S}&_{\text{uncertainty}} =  \mathbb{E}_{d\sim p_{\text{data}}(D)} 
    \mathcal{L}\left( d', d\right) 
    \label{eq:forward} 
\end{align}
\end{subequations} 
Next, we discuss how $S_{uncertainty}$ is used in \emph{uncertainty sampling} during data selection.

\subsection{Data Sampler}
\label{sec:select}

The process of data selection identifies $N$ data instances to be labeled such that the overall generation quality is improved.
This can be achieved by learning the structure over data $D$ \cite{tosh2018interactive}. We adopt a simple technique to represent each data instance $d$ as a bag-of-word (BOW) vector and further divide each \emph{attribute-type} (first layer) into $k$ clusters (\emph{sub-type}, second layer) with the K-means algorithm\footnote{Using the implemented version from \url{https://scikit-learn.org}} \cite{alsabti1997efficient}, which splits data instances into k clusters based on selected centroids. 

We first rank the order of batch-size samples within each \emph{sub-type} using \emph{uncertainty scores} (in \cref{sec:cyclic}) so that experts can annotate the ones with the least confident scores first.
At sampling time, we obtain unlabeled data instances from all \emph{sub-types} across all \emph{attribute-types} iteratively. 
One sample is obtained from each \emph{sub-type} before moving on to the next \emph{sub-type}.

For data presented to be labeled, the system also suggests labels in order to reduce annotation efforts. 
\emph{DART} employs a simple retrieval-based technique to obtain a text label $t$ for each data instance $d$.
Using the BOW representation of $d$, we simply find the most similar $d'$ (cosine similarity) in the labeled pool of $(d_i,t_i)$ pairs, and use its text label $t_i$ as the suggestion\footnote{There are no suggestions for the initial batch of annotations.}. 
The sampling process continues until either all data instances are labeled or a satisfactory 
threshold value is reached for the quality metric on the labeled corpus (as defined in \cref{sec:estimator}). 

\subsection{Quality Estimator}
\label{sec:estimator}

To better manage the annotation process, we include the diversity metrics used in \cite{constrained}: \emph{number of unique tokens}, \emph{number of unique trigrams}, \emph{Shannon token entropy}, \emph{conditional bigram entropy}.
Following \cite{novikova2016crowd}, we also measure various types of lexical richness including \emph{type-token ratio (TTR)} and \emph{Mean Segmental TTR (MSTTR)} \cite{lu2012relationship}, where higher values of TTR and MSTTR correspond to more diverse corpus.
\emph{DART} displays these scores on the \emph{Status Display} as shown in Figure~\ref{fig:anno}. 
These scores serve as on-the-fly quality estimates that help experts decide when sufficient labels have been collected.

\section{User Interface}
\label{sec:interface}

\begin{figure*}[h]
  \centering
\includegraphics[width=0.9\textwidth]{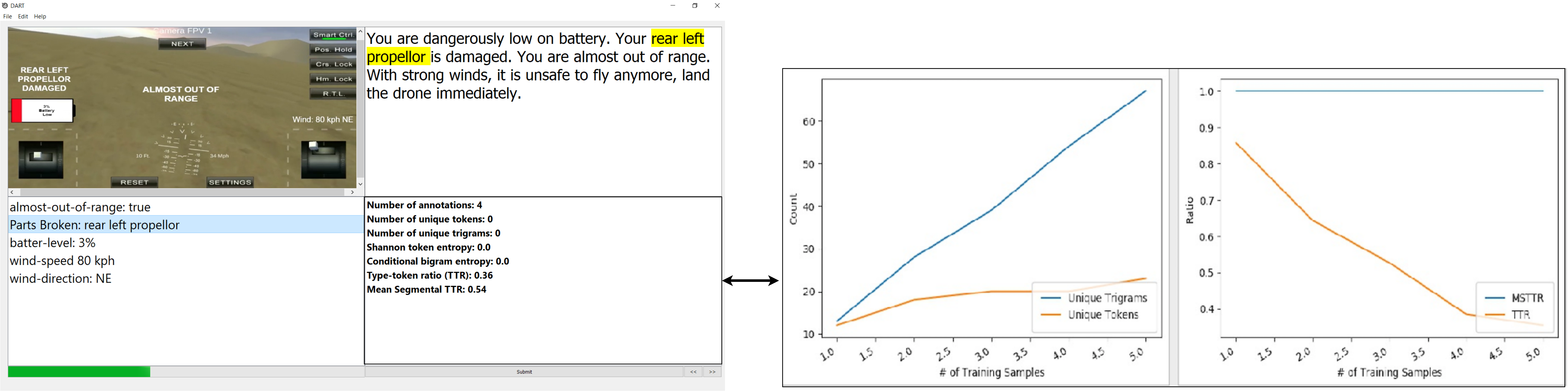}
\caption{ Screenshot of sample annotation interface showing the provided data, and the text box for annotation.
On the bottom right, user can select either the statistics or plots indicating the corpus diversity and annotation count.
}
\label{fig:anno}
\end{figure*}

\begin{figure*}[h]
  \centering
\includegraphics[width=0.7\textwidth]{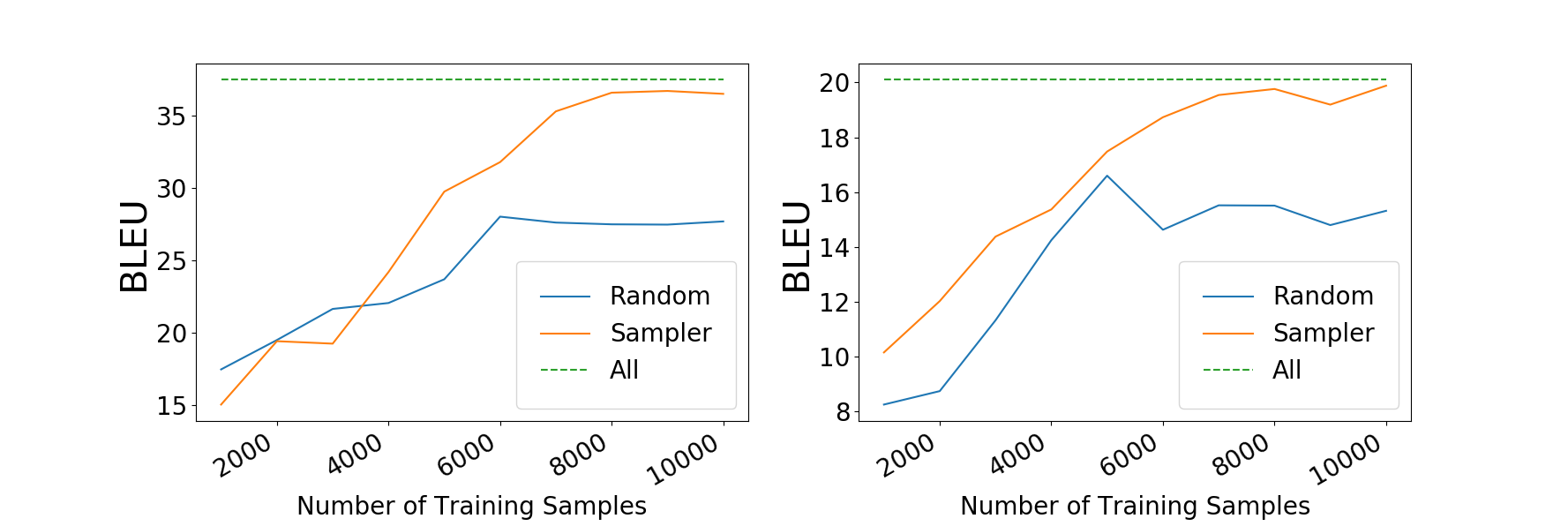}
\caption{ Performance comparison between DART's \textbf{data sampler}, \textbf{random sampling}, and retrieving labels from the full dataset (\textbf{ALL}) on E2E (Left) and the Weather (Right) datasets (42k data instances for E2E and 32k for weather), using the same retrieval method. }
\label{fig:compare}
\end{figure*}

The interactive interface divides the interactive application window into a few compartments: 
\emph{DART} includes a configuration editor interface that allows the experts to modify the delimiter (\emph{e.g.} ``,'') between \emph{attribute:value} pairs.
For graph structured input, the delimiter (\emph{e.g.} ``\_\_'') is used to identify the attribute tags instead. 
Note that the system supports three granularity of tokenization: (1) word, (2) character, and (3) byte-pair encoding (BPE) \cite{sennrich2016neural}.
The top half of Figure~\ref{fig:anno} shows the main annotation page where experts can input constructed sentences into the text boxes based on suggested texts and the provided image (or short clip)~\footnote{The use of pictures is shown to elicit better data \cite{novikova2016crowd}}.
As the expert annotates, the progress bar below the text box indicates when the background \emph{uncertainty scorer} training session will begin. 
The bottom half of Figure~\ref{fig:anno} shows the annotation progress statistics, including the percentage of data types that have been annotated and the quality of overall templates. 
When a specified number of annotations has been created, experts can download both the annotated data samples along with data with predicted labels. 
In general, a high-quality corpus maintain a high corpus diversity (\emph{e.g.} a MSTTR score of $0.75$ or TTR of $0.01$ in the E2E dataset \cite{novikova2016crowd}) even as the number of annotation increases.

\section{Experiments}
\label{sec:exp}

\paragraph{Data.}
\label{sec:data}

We use two different types of structured data: 
(A) Attribute-value pairs as used in the crowd-sourced \emph{E2E} dataset \cite{novikova2017e2e}, and
(B) the graph-structured data as defined in \cite{constrained} on the \emph{weather} domain.
To simulate the annotation process, we employ the given training, development and test sets of each datasets for annotation tool evaluation, with the test set kept fixed.
This amounts to roughly 42k samples for E2E and 32k for the weather training sets.

\comment{
We use the attribute-value pairs as used in the crowd-sourced \emph{E2E} dataset (42k samples) \cite{novikova2016crowd}.
To simulate the annotation process, we employ the given training, development and test sets of each datasets for annotation tool evaluation, with the test set kept fixed.}

\comment{

\begin{figure}[h]
  \centering
\includegraphics[width=0.4\columnwidth]{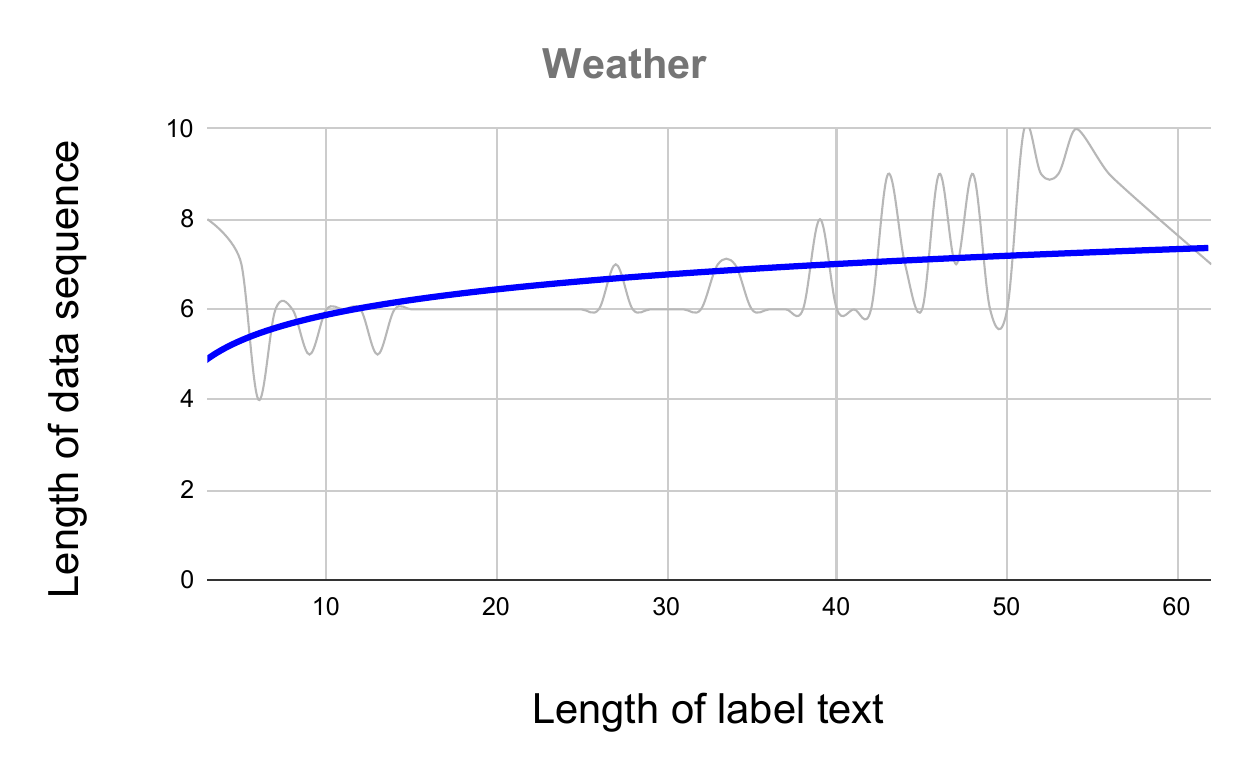}
\caption{ Log-linear trend-line of data length against label text length on the \emph{weather} dataset. }
\label{fig:weather_data}
\end{figure}
\vspace{-4mm}
\begin{figure}[h]
  \centering
\includegraphics[width=0.4\columnwidth]{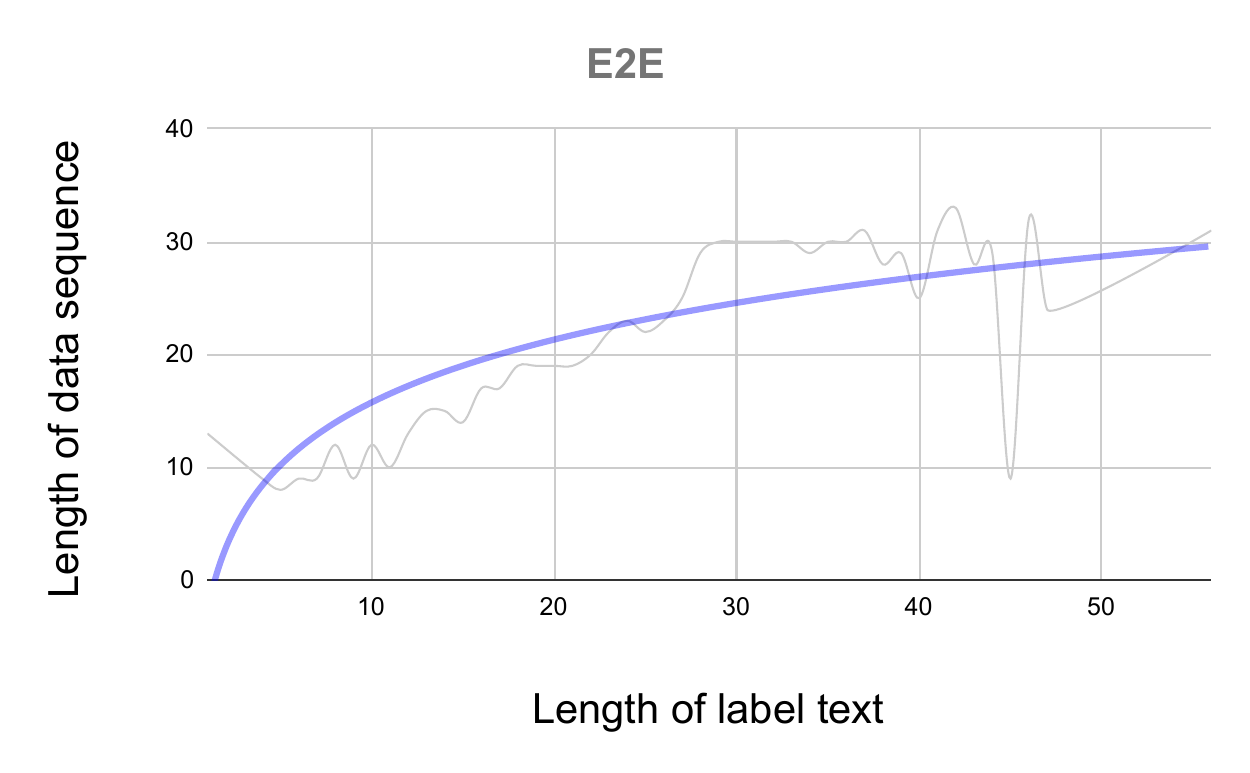}
\caption{ Log-linear trend-line of data length against label text length on the \emph{E2E} dataset. }
\label{fig:e2e_data}
\end{figure}

}


\paragraph{Simulation Study.}
To evaluate the effectiveness of \emph{DART}, we perform a simulated experiment for each of our two datasets, E2E and Weather. We simulate the labeling process using both the retrieval-based method (\textbf{Sampler}, as discussed in section \ref{sec:select}) and the baseline approach using random selection of data (\textbf{Random}) and compare the performance of the two methods relative to using the full dataset (\textbf{All}).

Results for the two datasets, E2E and Weather, are presented in Figure~\ref{fig:compare}. On both datasets, the data sampler allows the retriever to obtain the same performance (\emph{i.e.} with similar BLEU score as \emph{ALL}) using only 10k labeled data instances, which is significantly less than that of the original dataset (\emph{i.e.} 42k for \emph{E2E} and 32k for \emph{Weather}). 
As such, the number of required annotations to arrive at the same performance using all labeled data is significantly reduced for both datasets, to one-fifth of the original dataset size. In contrast, performance obtained using the \emph{Random} selection is significantly worse (on the order of 6 to 10 BLEU points lower than the baseline) compared to using the \emph{Sampler} selection while matching the number of training samples.

\comment{
To evaluate the effectiveness of \emph{DART}, we perform a simulated experiment on the E2E corpus. We simulate the labeling process using both the retrieval-based method (\textbf{Sampler} and the baseline approach using random selection of data (\textbf{Random}) and compare the performance of the two methods relative to using the full dataset (\textbf{All}).
Results on E2E are presented in Figure~\ref{fig:compare}. The data sampler allows the retriever to obtain the same performance (\emph{i.e.} with similar BLEU score as \emph{ALL}) using only 10k labeled data instances, which is significantly less than that of the original dataset (\emph{i.e.} 42k). 
As such, the number of required annotations to arrive at the same performance using all labeled data is significantly reduced for both datasets, to one-fifth of the original dataset size. In contrast, performance obtained using the \emph{Random} selection is significantly worse (on the order of 6 to 10 BLEU points lower than the baseline) compared to using the \emph{Sampler} selection while matching the number of training samples.}


\section{Conclusions}
\label{sec:conclusions}

While a wide range of annotation tools for NLP tasks exists, most of these tools are targeted at non-textual labels.
\emph{DART} is designed to enable the ease of annotation where the labels are textual descriptions and the inputs are structured data.
This is the initial version of the tool, and we hope to extend it to include a web-based version and to 
expand its functionality in the following ways:
(1) support different types of encoders, and
(2) improve upon the data sampling process.

\section*{Acknowledgements}
This research was funded in part by the German Research Foundation (DFG) as part of SFB 248 ``Foundations of Perspicuous Software Systems''. We sincerely thank the anonymous reviewers for their insightful comments that helped us to improve this paper.

\bibliographystyle{coling}
\bibliography{coling2020}

\end{document}